\documentclass[sigconf]{acmart}
\usepackage{graphicx}
\usepackage{subfigure}
\usepackage{multirow}
\newcommand{\bfsection}[1]{\vspace*{0.1cm}\noindent\textbf{#1.}}
\newcommand{\bfs}[1]{\vspace*{0.1cm}\noindent\textbf{#1}}

\AtBeginDocument{%
  \providecommand\BibTeX{{%
    \normalfont B\kern-0.5em{\scshape i\kern-0.25em b}\kern-0.8em\TeX}}}

\setcopyright{acmcopyright}
\copyrightyear{2021}
\acmYear{2021}
\acmDOI{10.1145/XXXXXX.XXXXXX}
\setcopyright{rightsretained}
\acmConference[MM '21] {Proceedings of the 29th ACM Int'l Conference on Multimedia}{October 20--24, 2021}{Chengdu, China.}
\acmBooktitle{Proceedings of the 29th ACM Int'l Conference on Multimedia (MM '21), Oct. 20--24, 2021, Chengdu, China}
\acmPrice{15.00}
\acmISBN{978-1-4503-8651-7/21/10}

\setcopyright{acmcopyright}
\copyrightyear{2021} 
\acmYear{2021} 
\setcopyright{acmcopyright}
\acmConference[MM '21]{Proceedings of the 29th ACM International Conference on Multimedia}{October 20--24, 2021}{Chengdu, China}
\acmBooktitle{Proceedings of the 29th ACM International Conference on Multimedia (MM '21), October 20--24, 2021, Chengdu, China}
\acmPrice{15.00}
\acmDOI{10.1145/3474085.3475439}
\acmISBN{978-1-4503-8651-7/21/10}

\begin{document}

\title{Dual Graph Convolutional Networks with Transformer and Curriculum Learning for Image Captioning}

\author{Xinzhi Dong}
\affiliation{%
  \institution{School of Computer, Wuhan University}
  \streetaddress{}
  \city{Wuhan}
  \state{Hubei}
  \country{China}
}
\email{dongxz97@whu.edu.cn}

\author{Chengjiang Long}
\authornote{This work was co-supervised by Chengjiang Long and Chunxia Xiao.}
\affiliation{%
  \institution{JD Finance America Corporation}
  \streetaddress{}
  \city{Mountain View}
  \state{CA}
  \country{USA}}
\email{cjfykx@gmail.com}

\author{Wenju Xu}
\affiliation{%
 \institution{OPPO US Research Center, InnoPeak Technology Inc.}
 \streetaddress{}
 \city{Palo Alto}
 \state{CA}
 \country{USA}}
\email{xuwenju123@gmail.com}

\author{Chunxia Xiao}
\authornote{Corresponding author.}
\affiliation{%
  \institution{School of Computer, Wuhan University}
  \streetaddress{}
  \city{Wuhan}
  \state{Hubei}
  \country{China}}
\email{cxxiao@whu.edu.cn}

\renewcommand{\shortauthors}{Trovato and Tobin, et al.}

\begin{abstract}
  Existing image captioning methods just focus on understanding the relationship between objects or instances in a single image, without exploring the contextual correlation existed among contextual image. In this paper, we propose Dual Graph Convolutional Networks (Dual-GCN) with transformer and curriculum learning for image captioning. In particular, we not only use an object-level GCN to capture the object to object spatial relation within a single image, but also adopt an image-level GCN to capture the feature information provided by similar images. With the well-designed Dual-GCN, we can make the linguistic transformer better understand the relationship between different objects in a single image and make full use of similar images as auxiliary information to generate a reasonable caption description for a single image. Meanwhile, with a cross-review strategy introduced to determine difficulty levels, we adopt curriculum learning as the training strategy to increase the robustness and generalization of our proposed model. We conduct extensive experiments on the large-scale MS COCO dataset, and the experimental results powerfully demonstrate that our proposed method outperforms recent state-of-the-art approaches. It achieves a BLEU-1 score of 82.2  and a BLEU-2 score of 67.6. Our source
code is available at {\em \color{magenta}{\url{https://github.com/Unbear430/DGCN-for-image-captioning}}}.
\end{abstract}




\maketitle

\begin{figure}[ht]
\begin{center}
\includegraphics[width=0.46\textwidth]{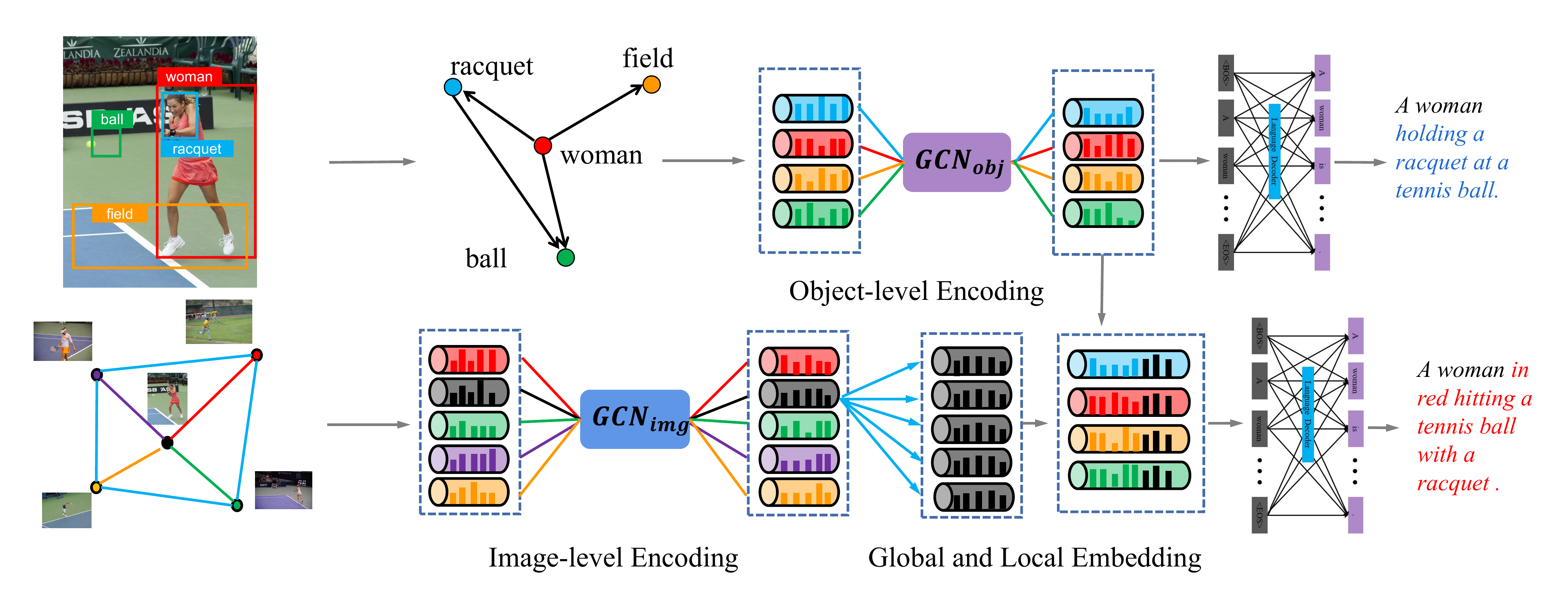}
\end{center}
\vspace{-0.1in} 
\caption{An illustration example of image captioning task, taking Graph Convolutional Networks (GCNs) as the visual encoder and a sequence-to-sequence model as a linguistic decoder. The top represents the visual encoder via a single object-level GCN on a single image, and the bottom indicates the visual encoder through the combination of both the object-level GCN and an image-level GCN. The visual embedding feature extracted in either way can be fed into a linguistic decoder to generate a reasonable text description.}
\vspace{-0.1in} 
\label{fig:intro}
\end{figure}

\section{Introduction}
Image captioning aims to generate a syntactic and correct description for a given single image, {\em a.k.a.}, a translation from image to language. One example is illustrated in Figure~\ref{fig:intro}. 
It requires a clear understanding of different objects and the relation between them, and the grammar in the caption. Otherwise, the model will miss some objects or generate corrections that do not conform to common sense. Many deep learning-based methods and algorithms have been proposed for image captioning~\cite{2017Long,Johnson2016DenseCap,2017Faster,2016Show}. They heavily rely on Convolutional Neural Networks (CNNs)~\cite{Yao2020Hierarchy, Long:CVPRW2017, Long:CVPRW2019A, Zhang:ICME2020, Zhang:TCSVT2021, Islam:AAAI2021} to extract hierarchical image features and take Recurrent Neural Networks (RNNs) like long-short-term memory (LSTM)~\cite{cornia2019show,Karpathy2016Deep,Vinyals2015Show,xu2015show, Long:WACV2019} as the language model to generate descriptive captions. However, these CNN-RNN based methods suffer from the limitations of fixed captions. Therefore, captions generated by these methods are far from being able to exactly describe the content of the input image.

Graph Convolutional Networks (GCNs)~\cite{2018Exploring, Yao2020Hierarchy, 2018Auto, Shi:CVPR2021} are then exploited on the structured graphs to enrich region representations for image captioning tasks. Recently, transformer~\cite{2017Attention, 2018BERT} has shown promising performance when dealing with serialization information. What's more, transformer has been recognized as the state of the art in sequence modeling tasks like language understanding and machine translation. Although there exists some efforts~\cite{cornia2020m2,he2020image} of applying transformer to solve the image captioning task and achieve some success, the potential of integrating GCNs with transformer is still largely under-explored.

\begin{figure*}[h]
\vspace{-0.20cm}
\begin{center}
\hspace{-0.23cm}\includegraphics[width=0.95\linewidth]{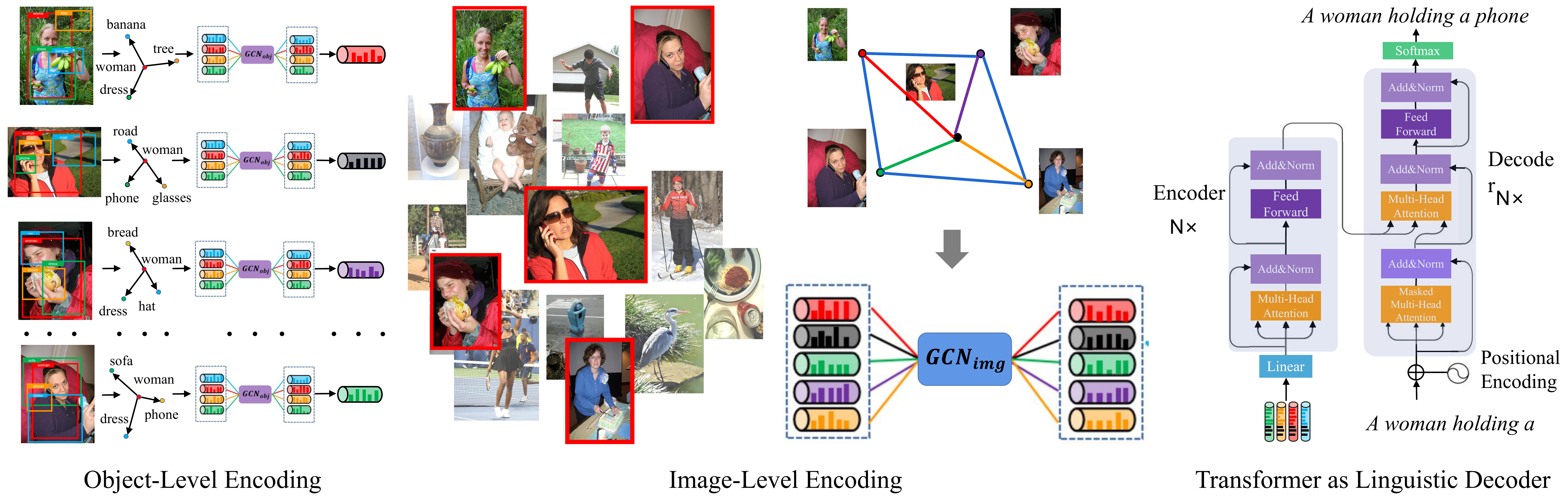}
\end{center}
        \vspace{-0.1in}  
   \caption{The overview of our proposed framework for image captioning. It consists of three main parts: an object-level GCN, an image-level GCN, and the transformer as the linguistic decoder.  In specific, given an input image, we use an object-level GCN and an image-level GCN to re-encode both the object-level and image-level features. 
With a combination of 
visual embedding features as the input, we take transformer as the linguistic decoder to generate a reasonable text description. 
}
\label{fig:overview}
\end{figure*}

In this paper, we propose a novel image captioning framework taking Dual Graph Convolutional networks (Dual-GCN) as the visual encoder and take the transformer as the linguistic decoder. As shown in Figure~\ref{fig:overview}, our framework integrates an object-level GCN and an image-level GCN for visual feature encoding. The first GCN focuses on objects in local regions to explore spatial object-level relation~\cite{2017Detecting,2017The,2010Grouplet,0Deep}, while the second one targets on the image-level similarity relation~\cite{2017Few,2011Image} amongst multiple similar images. We calculate the similarity of different images and select the images with high similarity as a meaningful complementary global visual embedding to ensure a more reasonable and accurate text description generation. In this way, our framework can learn the global and local visual embedding with the well-designed Dual-GCN framework. This significantly distinguishes our method from other GCN-based models \cite{2018Exploring, Yao2020Hierarchy}, which only considers the local object regions.


Unlike most existed approaches taking RNNs as the linguistic decoders, we choose transformer as our linguistic decoder instead. 
Due to the well-designed multi-head self-attentions, the transformer has the advantage of parallelization and has a stronger ability to extract visual features. It shows superior performance in the fusion of visual features and text features. 
Therefore, taking in the global and local visual embedding as input, our transformer linguistic decoder can generate a syntactic and correct description for the single input image.


It is worth mentioning that existing caption models~\cite{xu2015show, Vinyals2015Show,  2016Boosting, 2016Image, 2017Long, yao2017incorporating, 2018Bottom, 2018Auto, 2018Exploring, 2019Attention,  2019Attention,  Yao2020Hierarchy, cornia2020m2} are trained in a straightforward manner, in which all the training data are fed into the model evenly and equally, ignoring the fact that different training images may contain information at different levels. We argue that a difficult image capturing a complex scene usually needs a long caption  to describe the objects and their spatial positioning. Therefore, an under-trained model could be misled by the wrong gradients when trained on difficult images. Inspired by the human nature learning process in an easy-to-hard manner, we adopt the curriculum learning~\cite{xu-etal-2020-curriculum} as a training strategy. 
A cross-review mechanism is then introduced to distinguish the difficulty of the training images. In particular, we partition the entire training dataset evenly into multiple subsets and train an image caption model on each subset. For any training example, its difficulty level is determined across all the other learned models, each of which never takes it as the training example. With the difficulty levels assigned on all the training examples, we can train our proposed model on easy images at the beginning. Then we keep adding difficult images as the training examples. Obviously, the newly added information dramatically enhances our model's robustness and generalization ability in terms of understanding different objects and the relation between them.
To summarize, the main contributions of this paper are three-fold:
\begin{itemize}
\item We propose a novel framework that takes Dual-GCN as a powerful visual encoder and  transformer as the linguistic decoder for image captioning. The first object-level GCN captures the region to region relation in a single image, and the second image-level GCN captures the image's similarity relation. This treatment enables both global and local visual information well encoded and fed into the transformer to generate better captioning descriptions.
\item To our best knowledge, we are the first one to apply the Curriculum Learning as the training strategy on image captioning model in an easy-to-hard manner. Note that the difficulty level is determined by our well-designed cross-review mechanism.
\item We conduct extensive experiments 
on the MS COCO dataset. The results strongly demonstrate that our proposed approach outperforms state-of-the-art methods. It achieves a BLEU-1 score of 82.2 and a BLEU-2 score of 67.6. 
\end{itemize}

\section{Related work}
%
{\bf Image Captioning} is a cross-modal task combining computer vision and natural language processing. It has been widely studied as a deep learning task ~\cite{2012ImageNet} in recent years. The dominant paradigm in modern image captioning are sequence learning methods~\cite{2017Long,Vinyals2015Show,xu2015show,yao2017incorporating,2016Boosting,2016Image}. Vinyals et al.~\cite{Vinyals2015Show} used a deep CNN to extract the visual features of the image and adopted LSTM \cite{2018Exploring} as the language model to guide the language sequence generation. Instead of learning a visual attention map on the input image, ~\cite{2017Knowing} proposed an encoder-decoder model that learns an adaptive attention map to combine the visual feature and language embedding. ~\cite{2018Bottom} proposed bottom-up attention and a top-down attention mechanism for caption generation. Chen et al.~\cite{2016SCA} extended the spatial -wise and channel-wise attention mechanism to extract both spatial features and semantic features from an image. Unlike these mentioned models, Yao et al.~\cite{2018Exploring} adopted the graph neural network to integrate spatial and semantic relationships between different objects. In recent years, transformer-based methods~\cite{2018Auto} show great potential in image captioning. Herdade et al.~\cite{herdade2019image} used the transformer architecture for image captioning. Then Li et al.~\cite{li2019entangled} proposed to take an external tagger providing visual information and semantic knowledge to enhance the performance of the transformer model. As reverse process of image captioning, Hu et al.~\cite{2021A}~\cite{Hu:Arxiv2021} used diverse conditional GAN \cite{xu2019adversarially} for text representation and generated video animation from single still image ~\cite{HU2020102812}. Fang et al.~\cite{2020A} solved a series of problems of analyzing complex text and finally generating corresponding images, and also used scene graph and generates more realistic images~\cite{2018Narrative}.   

{\bf Graph Neural Networks}
have been widely used in many applications to build up the relations between different features, such as nature language processing and computer vision tasks~\cite{Long:ICCV2015, Long:IJCV2016, 2016Semi, Long:CVPR2017, 2018Exploring, Hua:TPAMI2018}. In particular, ~\cite{2016Semi} solves the computing bottleneck by learning the Laplacian polynomials of graphs. When combining with the LSTM~\cite{2018Exploring}, GCN can be used for image captioning task. It is used to extract relations between different objects, which achieved great results in the final caption generation. However, these GCN based methods only explore the object relations within one single image. Instead, we propose a Dual-GCN module to aggregate information extracted from both the input image and other similar images. 

{\bf Transformer} is first proposed by~\cite{2017Attention}. It is a combination of multi-layer encoder and decoder. The encoder consists of a stack of self-attention and feedforward layers, while the decoder uses self-attention on the text and the cross-attention to the last encoder layer. Herdade et.al ~\cite{herdade2019image} then used the transformer in image captioning. They combined geometric relationships between different input objects. ~\cite{cornia2020m2} used a priori knowledge module to improve the transformer encoder and also used mesh connection between multi-layer encoder layer and decoder layer to improve the hierarchical diversity of feature extraction. Li et al. ~\cite{li2019entangled} used the transformer to make use of visual information and additional semantic knowledge given by an external tagger. In this paper, we take transformer as the linguistic model to generate a reasonable text description based on the extracted visual embedding feature.

{\bf Curriculum Learning} is first proposed by ~\cite{10.1145/1553374.1553380} in the machine learning area. This method first classifies the sub-dataset according to the difficulty, such as finding examples that are easy to implement in the data set. Then it sets up an easy-to-hard curriculum for the learning procedure. Such a human-like learning strategy considerably improves the model's capability. 
Curriculum learning has been widely used in the computer vision area ~\cite{2018MentorNet,2018CurriculumNet, Xiao:TMM2020}. With the continuous expansion of transformer in computer vision, we propose to use a curriculum learning strategy aiming at improving our Dual-GCN model's performance on image captioning task.

\section{Proposed Method}
Our method aims to generate a caption that correctly describes the content of an input image. It comprises three main parts: Dual-GCN encoder, transformer decoder, and curriculum learning strategy. In this section, we will introduce each framework and our algorithm in detail. Figure \ref{fig:overview} shows our main framework.

Given an image $I$, we take a textual caption to describe its content, denoted by $S$. First, we take the Faster-RCNN~\cite{2017Faster} as a backbone to produce a set of detected objects $V_{obj}=\{v_i\}^O_{i=1}$ where $O$ denotes different object regions in an image. $v_i$ denotes a $C$-dimensional feature representation of the $i$-th object region. 

\subsection{Dual-GCN Encoder}
To learn a better latent representation of the input image, we propose a Dual-GCN encoder, which consists of an object-level GCN and a image-level GCN. Based on the region feature from the Faster-RCNN detector, our Dual-GCN encoder produces better features by seeking the spatial relation within the same image and the similarity among different images.


\bfsection{Object-Level GCN}
We take the object-level GCN to build a spatial relation between different objects within the same image. To this end, we consider each feature extracted from one single object region as a vertex and consider the relative position amongst different objects as a directed line segment. Specifically, we build a spatial graph $G_{obj} =(V_{obj},E_{obj})$, where $E_{obj}=\{(v_i, v_j)\}$ is the set of spatial relation edges between region vertices, and the edge $(v_i,v_j)$ represents the relative geometry relationship between the $i$-th and the $j$-th objects. More specifically, following \cite{2018Exploring}, we use the IOU, relative distance, and the relative angle between two objects as the relative geometry relationship to determine the edges, these relationships can be divided into "inside", "cover", "overlap" and "no" relation" and so on. Therefore, each vertex $v_i$ is encoded via a modified GCN as:
\begin{equation}
    v^{obj}_i=\sigma(\sum_{v_j\in N(v_i)}W_{v_j\rightarrow v_i}v_j+b_{v_j\rightarrow v_i)}),
\end{equation}
where each $v_i$ is a real-valued vector and it represents the information of the nodes in the graph. $\sigma$ is an activation function, and $N(v_i)$ denotes a set of neighbor vertices of $v_i$.  $v_j\rightarrow v_i$ indicates the direction from $v_j$ to $v_i$. 
$W_{v_j\rightarrow v_i}$ and $b_{v_j\rightarrow v_i}$ denote the transformation matrix and the bias vector, respectively.

\bfsection{Image-Level GCN} 
The information extracted from a single image is very limited. As a remedy, we propose a novel image-level GCN, which enhances the information flow in the graph by considering other relevant images. In image feature extraction, we observe that images sharing a certain degree of similarity to the input image can be an auxiliary information source for the input image captioning.


Let ${\bar v}_j$ stands for the latent representation of $j$-th image, which contains a set of objects $V^j_{obj}=\{v^{obj}_i\}^O_{i=1}$. We define:
 \begin{equation}
     {\bar v}_j =\frac{1}{O}\sum_{i=1}^{O}v^{obj}_i,
\end{equation}

Based on the latent representation, we seek similar images ${\bar v}_s \in \Omega({\bar v}_j)$ according to the $\mathcal{L}_2$ distance between the features of the input image and other images from the same dataset. This is formulated as:
 \begin{equation}
    {\bar v}_s \in \Omega({\bar v}_j):=\left[\sum_{t=1}^{C}{({\bar v}^t_s-{\bar v}^t_j)^2}\right]_K,
\end{equation}
where $C$ is the dimension of the feature; $[\cdot]_K$ is to select $K$ images with lowest distance score. Once we obtain the top-K images, 
we establish an image-level graph $G_{img} =(V_{img},E_{img})$ by considering the image feature as a node $V_{img}$ and the similarity between paired images as an edge $E_{img}$. 
Then we encode the features represented by each image node in the graph:
\begin{equation}
     u^{img}_j=\sigma(\sum_{{\bar v}_s\in \Omega({\bar v}_j)}W{\bar v}_s+b),
\end{equation}
where $\Omega({\bar v}_j)$ denotes the set of similar images of ${\bar v}_j$. Note that in our image-level graph $u^{img}_j$ represents the feature of an image and its $K$ nearest neighbor images.

\bfsection{Global and Local Visual Embedding} 
Since we only need to generate the caption of a given image, we choose to extract the feature vector of GCN encoded in the node of the original image. Then we concatenate the feature vectors of GCN recoded on image-level similarity with the original image encoded on object-level as the input of a transformer as the language model for text prediction:
\begin{equation}
\begin{split}
 U_j &= Concat(V^j_{obj},v^{img}_j) \\
 &= [concat(v_1^{obj}, u^{img}_j), \ldots, concat(v_O^{obj}, u^{img}_j)].   
\end{split}
\end{equation}

In this way, we combine the information extracted from the input image and other similar images. The final visual embedding representation integrates the characteristics of each object in the input image locally and the contextual object information from multiple similar images globally. Then, the global and local visual embedding information is fed into the linguistic decoder to ensure a reasonable caption generation.

\subsection{Transformer as Linguistic Decoder}
As shown in Figure \ref{fig:overview}, after we get the global and local embedding $U = \{U_j| j=1, \ldots, O\}$ of the input image, we take a transformer decoder to generate a caption correctly describing the content of the image.



\bfsection{Encoding Layer} The encoding layer is based on multi-head attention layer, which can be formulated as: 
\begin{equation}
     U_{att}=softmax(\frac {W_qU(W_kU)^T} {\sqrt {d}}) W_vU,
\end{equation}
where $W_q$, $W_k$ and $W_v$ are the learnable parameters used to create the query, keys and values respectively; $d$ is a constant used for normalization. Besides, a residual structure is adopted to stabilize the attention module. This ensures that the transformer can extract the spatial information of different objects to a greater extent in the self-attention part. Thus, we achieve:
\begin{equation}
    \bar U=AddNorm(U \oplus U_{att}),
\end{equation}
where $\oplus$ is an element-wise addition operation. Based on the above structure, multiple encoding layers are stacked in sequence so that each encoder layer receives the output calculated by the previous encoder layer. This is equivalent to creating a multi-layer high-level coding of the relationship between image regions, in which the higher coding layer can utilize and refine the previously recognized relationship and finally produce the output.




\bfsection{Decoding Layer}
We take a similar multi-head attention module to process text and image feature.
The decoder layer also contains feed-forward layer, and all components are encapsulated in addnorm operator. The structure of decoder layer can be described as:
\begin{equation}
    Q=AddNorm(\bar U,\bf Y),    
\end{equation}
where $\bf Y$ is the input sequence.
Like the encoder layer, our decoder layer takes a similar self-attention and residual structure. And multiple decoder layers are also stacked in sequence. The purpose is to extract higher-level text and image features and generate more accurate word distribution probability:
\begin{equation}
    \bar Q=AddNorm(Q \oplus Q_{att}),
\end{equation}
where $Q_{att}$ stands for the output of a multi-head attention module in the decoding layer. Finally, by inputting the generated results into the softmax layer, the word probability distribution of the currently generated words is obtained:

\begin{equation}
    P=Softmax (\bar Q).
\end{equation}

After getting the word probability distribution through the transformer, we get the probability distribution $P(S'|S,I)$ for the given image. Where $S$ is the ground-truth caption embedding matrix, $I$ is the image feature we used, and $S'$ is the currently generated captions. Through the algorithm of conditional probability, our model gets the whole caption probability for the given image.
  

 \begin{figure}[t!]
\begin{center}
\includegraphics[width=1.0\linewidth]{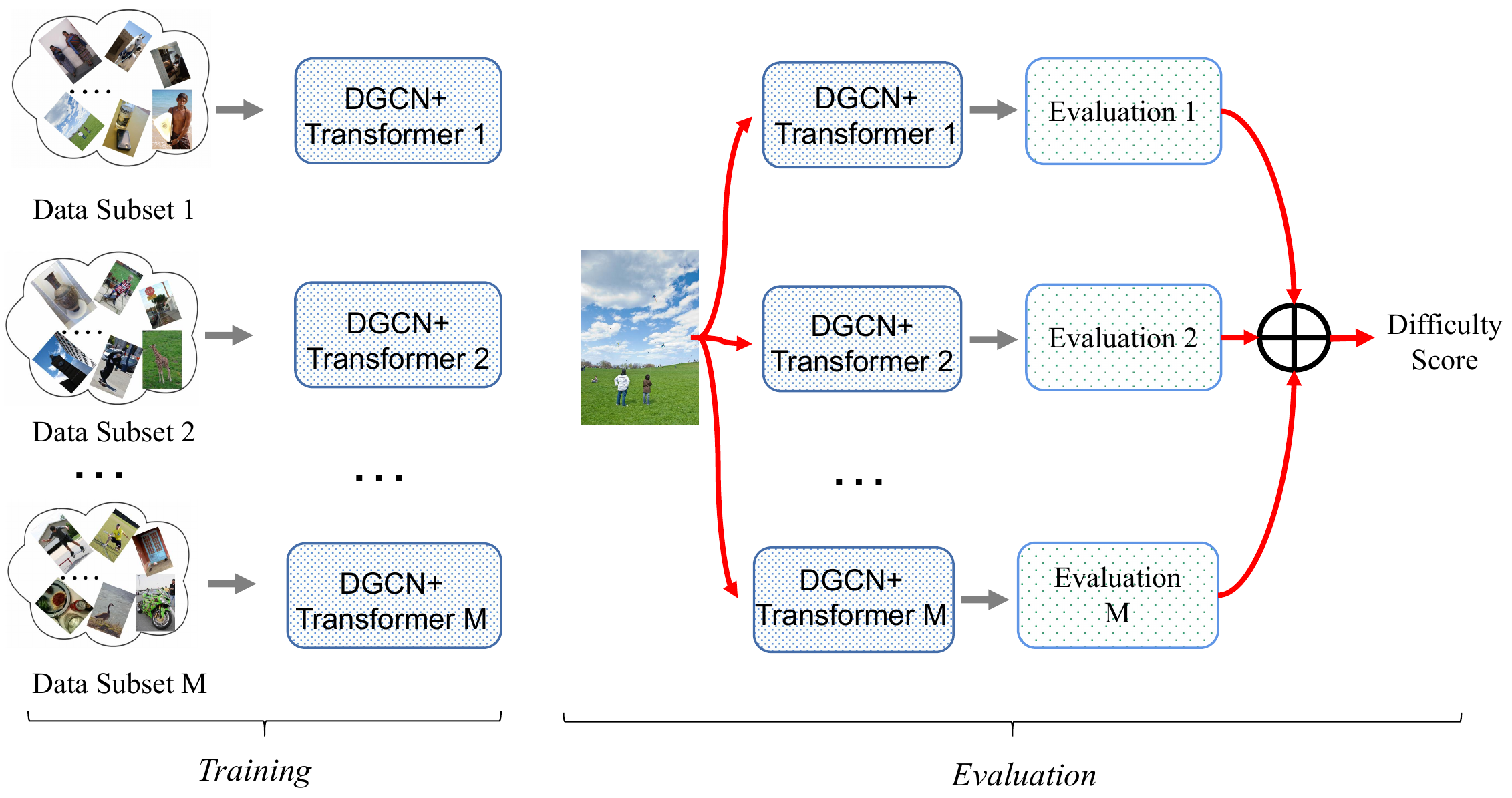}
\end{center}
\vspace{-0.12in} 
\caption{The cross-review mechanism used to determine the difficulty level for training examples. Inspired by ~\cite{xu-etal-2020-curriculum}, the dataset  is evenly divided into $M$ subsets. Then a Dual-GCN+Transformer model is trained on each data subset for caption generation. Given a training sample from the $i$-th data subset, we use all other $M-1$ trained Dual-GCN+Transformer models together except the $i$-th model to determine its difficulty level. Note that we use BLEU as the metric to calculate the corresponding difficulty score.}
\vspace{-0.2in}
\label{fig:cross-review}
\end{figure}
 
\subsection{Curriculum Learning}
To adopt the curriculum learning for our image captioning task, we have to determine the difficulty level from easy to hard.

\bfsection{Objective Function}
Let $D_{data}$ be the MSCOCO training dataset, and $\Theta$ be our Dual-GCN+Transformer caption model. In order to obtain a more comparable and stable difficulty score, we first uniformly split our training dataset $D_{data}$ into $M$ parts. Each sub-dataset is denoted by ${D_i}$ $(i=1,2,3...M)$. Then we train our Dual-GCN+Transformer caption models $\Theta_i$ on them respectively. Each model can only use $1/K$ of the training dataset for training. 
The parameters of these models can be learned by the following optimization:
\begin{equation}
    \Theta_i=\mathop{\arg\min}_{\Theta_i}\sum_{(I, S)\in{D_i}}\mathcal{L}_{ce}(\Theta_i(I),S),
\end{equation}
where $(I, S)$ is the image-to-text pair in the $i$-th traning dataset. $\mathcal{L}_{ce}$ defines the cross entropy loss function between the probability prediction by $\Theta_i(I)$ and the corresponding ground-truth caption $S$.

\bfsection{Difficulty Assessment} 
Most of the traditional ideas are based on word frequency. The difficulty is defined according to the number of times each word appears in the dataset. For example, in the MSCOCO training set, ``person'' appeared in 64,115 images and had 262,465 tag boxes, while ``bus'' only appeared in 3,952 images and had 6,069 tag boxes. Obviously, ``person'' is more accessible to generate than ``bus''.	However, since image caption is a cross-modal matching task from image to text, the difficulty evaluation should be consistent with the corresponding evaluation metrics like BLEU-1 and BLEU-2. Inspired by~\cite{xu-etal-2020-curriculum}, we adopt a cross review mechanism to determine the difficulty level for all the training examples, as shown in Figure~\ref{fig:cross-review}. 

After we train our caption models on $M$ sub-datasets, we evaluate the difficulty level for each training example. Note that each image-to-text example $(I, S) \in D_i$ has been seen by model  $\Theta_i$ during training. Hence we use another model $\Theta_k$ to evaluate the difficulty of $(I, S)$:
\begin{equation}
\mathcal{E}_k(I,S) = 1-Metric(\Theta_k(I), S),
\end{equation}
where $\mathcal{E}_k(I,S)$ is the difficulty score of the text-image pair $(I, S)$. $Metric$ stands for a formula, which could be one of the image captioning metrics such as BLEU-1,BLEU-2,BLEU-3 and BLEU-4 ~\cite{2002A}. 
In this paper, we take the average of BLEU-1 score to indicate the difficulty. This is formulated as the sum of the evaluation scores of other caption models:
\begin{equation}
    DS((I, S)) =\frac{1}{M-1}\sum_{(I, t) \in D_i, k\neq i}\mathcal{E}_k(I,S),
\end{equation}
where $DS((I, S))$ is the difficulty score of the text-image pair (I,S) in sub dataset $D_i$. 


\bfsection{ Curriculum Learning Arrangement Part}
We first sort all text-image pairs according to their difficulty scores $DS$, and then divide them into $M$ sub datasets: $U_i$. The training dataset is then arranged from $U_1$ (the easiest) to $U_M$ (the hardest). The numbers of example in each categories are defined as $|U_1|, |U_2|, \ldots, |U_M|$. We take $M$ stages to train our model, it can be defined as $C_i$ $(i=1, \ldots, M)$. Notice that at each stage $C_i$, the image-to-text example is still shuffled to keep local stochastics, and each example from different stages does not overlap in order to prevent overfitting. For each learning stage $C_i$, we choose example according to a certain proportion from the aforementioned categories according to difficulty. The specific quantity is as follows:
\begin{equation}
\frac{|U_1|}{M}:\frac{|U_2|}{M}:....\frac{|U_M|}{M}.
\end{equation}
When the training stages are reached on $C_M$, we think the model should be ready to train on the whole MS COCO dataset. So we add another learning stage $C_{M+1}$, which is the entire MS COCO dataset.

\begin{table}[t!]
   \caption{Visualization examples of images with different difficulty levels. We can observe that the difficulty level increases with more and more complex background.}
         \vspace{-0.1in}  
   \begin{center}
    \setlength{\tabcolsep}{1.4mm}
    \begin{tabular}{p{1.3cm} |p{5.5cm}| p{0.8cm}}
       \hline
       {\quad Image} $I$ & {\quad\quad\quad\quad Ground-truth Caption $S$} & $DS(I,S)$ \\
       \hline
       \multirow{5}{*}{
       \begin{minipage}{0.1\textwidth}
       \includegraphics[width=1.3cm,height=1.2cm]{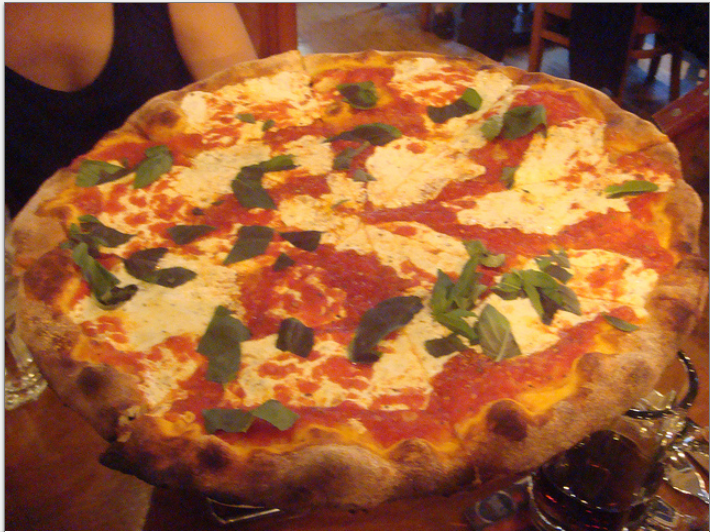} 
       \end{minipage}} & {\em\footnotesize A cheesy pizza siting on top of a table.} & \multirow{5}{*}{0.214} \\
       & {\em\footnotesize This large pizza has a lot cheese and tomato sauce.} &\\
       & {\em\footnotesize A prepared pizza being served at a table.} &\\
       & {\em\footnotesize A pizza with olive, cheese, tomato and onion.} &\\
       & {\em\footnotesize Baked pizza with herbs displayed on serving tray at table.}& \\
       \hline
       \multirow{5}{*}{
       \begin{minipage}{0.1\textwidth}
       \includegraphics[width=1.3cm,height=1.2cm]{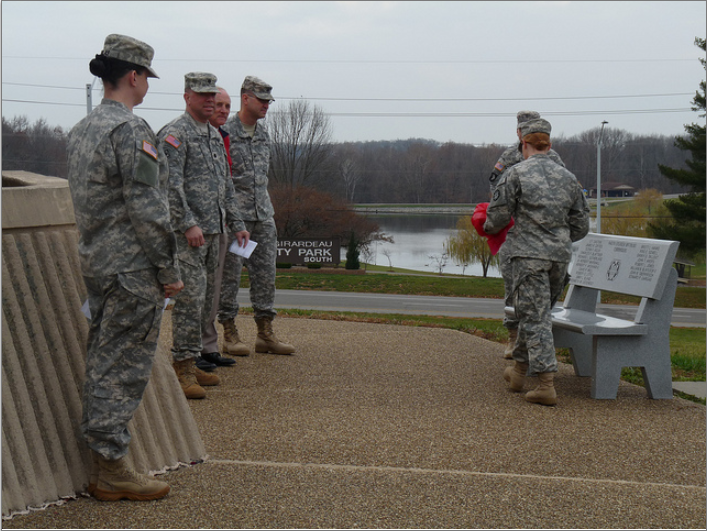} 
       \end{minipage}} & {\em\footnotesize A group of Army soldiers visit a memorial. } & \multirow{5}{*}{0.268} \\
        & {\em\footnotesize A group of soldiers standing next to a bench.} &\\
       & {\em\footnotesize Soldiers performing a ceremony in a city park.} &\\
       & {\em\footnotesize Members of the military are in a park.} &\\
       & {\em\footnotesize Military officers standing on a public park in uniform.}& \\
       \hline
       \multirow{5}{*}{
        \begin{minipage}{0.1\textwidth}
       \includegraphics[width=1.3cm,height=1.0cm]{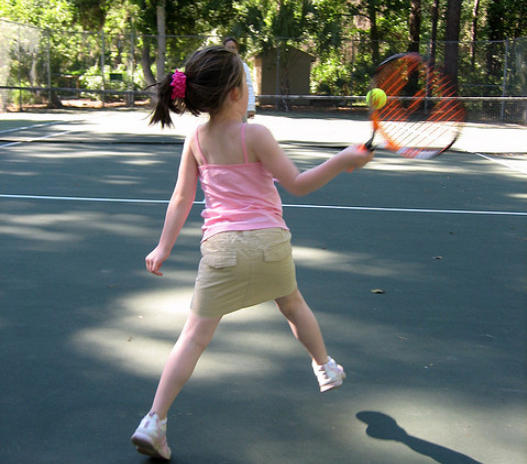} 
       \end{minipage}} & {\em\footnotesize Alittle in a pink shirt and khaki skirt playing tennis. } & \multirow{5}{*}{0.308} \\
        & {\em\footnotesize A young girl playing with her friend in a game of tennis.} &\\
       & {\em\footnotesize A girl is hitting a ball with a tennis racket.} &\\
       & {\em\footnotesize A small girl in action on a court with a racket.} &\\
       & {\em\footnotesize Military officers standing on a public park in uniform.}& \\
       \hline
   \end{tabular}
   \label{tab:N}   
   \end{center}
        \vspace{-0.2in}   
\end{table}




\section{Experiments}
\subsection{Datasets and Metrics}
\bfs{Microsoft COCO Dataset}~\cite{2014Microsoft} (MS COCO) is the dataset most commonly used in image captioning tasks. This dataset consists of about 82,700 training images and 40,500 validation images. There are five human-annotated descriptions for each image. Following the settings used by ~\cite{Karpathy2016Deep}, we take 5,000 images for validation, 5,000 images for testing and almost 110,000 images for training.
\begin{table*}
   \vspace{-0.3cm}
    \centering
     \setlength{\tabcolsep}{5mm}
    \caption{Comparison with the state of the art on the Microsoft COCO dataset, in single-model setting.}
    \label{tab:comp_with_state_of_the_art}     
       \vspace{-0.1in}      
    \begin{tabular}{c|c|c|c|c|c|c|c}
    \hline
    & {\small BLEU}-1 &{\small BLEU}-2 & {\small BLEU}-3 & {\small BLEU}-4 & {\small METEOR} & {\small ROUGE} & {\small CIDEr}  \\
    \hline
    Up-down~\cite{2018Bottom}  &  80.2 & 64.1 & 49.1  & 36.9   & 27.6   & 57.1  & 117.9 \\
    AOA-NET ~\cite{2019Attention}   &  81.0  & 65.8   & 51.4  & 39.4  & 29.1   & 58.9  & 126.9 \\
    \hline
    GCN-LSTM~\cite{2018Exploring}  &  80.8 & 65.5  & 50.8 & 38.7   & 28.5  & 58.5  & 125.3 \\
    GCN-LSTM+HIP~\cite{Yao2020Hierarchy} & 81.6  & 66.2 & 51.5 & 39.3 & 28.8  & 59.0 & 127.9\\
    SGAE  ~\cite{2018Auto} &  81.0  & 65.6   & 50.7 & 38.5  & 28.2 &58.6 & 123.8 \\
    $M^2$ Transformer~\cite{cornia2020m2} & 81.6 &  66.4 & 51.8  & {\bf 39.7}  & 29.4  & 59.2 & 127.9 \\
    \hline
    Dual-GCN+Transformer+CL     & {\bf 82.2}   & {\bf 67.6}  & {\bf 52.4}  & {\bf 39.7}   & {\bf 29.7}   & {\bf 59.7} & {\bf 129.2} \\
    \hline
    \end{tabular}
        \vspace{-0.2in}    
\end{table*}


\begin{figure*}[htbp]
\vspace{0.5cm}
\begin{center}
\hspace{-5.5cm}\includegraphics[width=2.75cm, height=2.0cm]{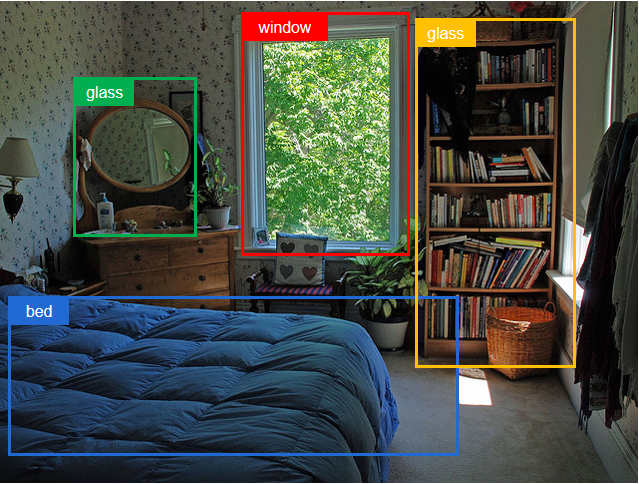} 
\includegraphics[width=2.75cm, height=2.0cm]{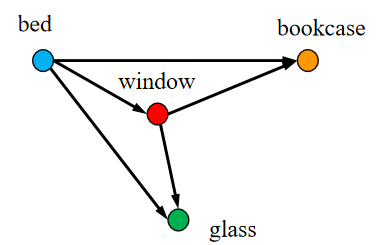} 
\hspace{0.25cm}\includegraphics[width=2.75cm, height=2.0cm]{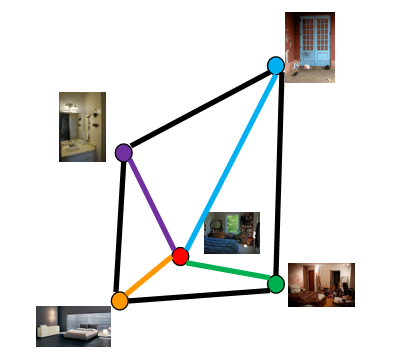} 
\begin{picture}(100,60)
\put(10,48){{\em\footnotesize [Up-down] Bedroom scene with a bookcase and a window.}}
\put(10,24){{\em\footnotesize [GCN-LSTM] A bed room with a large bookcase near a window.}}
\put(10,36){{\em\footnotesize [AOA-net] A bed room with a bookshelf full of books.}}
\put(10,12){{\em\footnotesize [M$^2$ Transformer] A bed room with a neatly bed and a book shelf.}}
\put(10,0){{\em\footnotesize [Dual-GCN+Trans+CL] A room has a bed with \textcolor{red} {blue sheets and a large bookcase.}}}
\end{picture}
\\
\vspace{0.05cm}
\hspace{-5.5cm}\includegraphics[width=2.75cm, height=2.0cm]{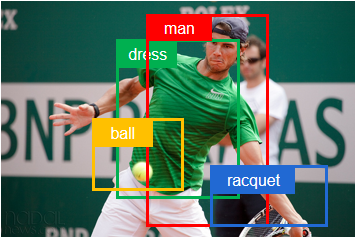} 
\includegraphics[width=2.75cm, height=2.0cm]{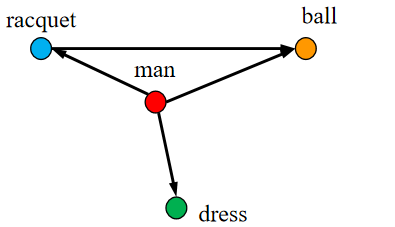} 
\hspace{0.25cm}\includegraphics[width=2.75cm, height=2.0cm]{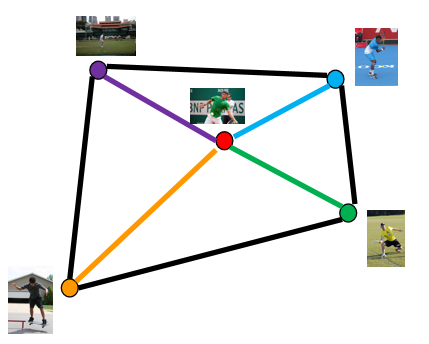} 
\begin{picture}(100,60)
\put(10,48){{\em\footnotesize [Up-down] A tennis player hits the tennis ball.}}
\put(10,24){{\em\footnotesize [GCN-LSTM] A tennis player is hitting a tennis basll.}}
\put(10,36){{\em\footnotesize [AOA-net] A picture of a tennis player about to hit a ball.}}
\put(10,12){{\em\footnotesize [M$^2$ Transformer] A tennis player is swinging at a tennis ball.}}
\put(10,0){{\em\footnotesize [Dual-GCN+Trans+CL] A man \textcolor{red} {holding a tennis racquet} in front of a tennis ball.}}
\end{picture}
\\
\vspace{0.05cm}
\hspace{-5.5cm}\includegraphics[width=2.75cm, height=2.0cm]{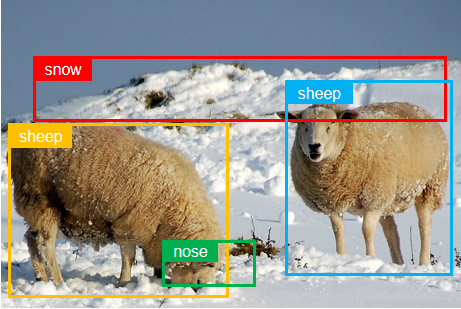} 
\includegraphics[width=2.75cm, height=2.0cm]{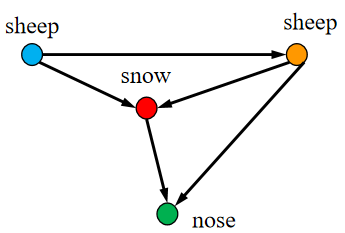} 
\hspace{0.25cm}\includegraphics[width=2.75cm, height=2.0cm]{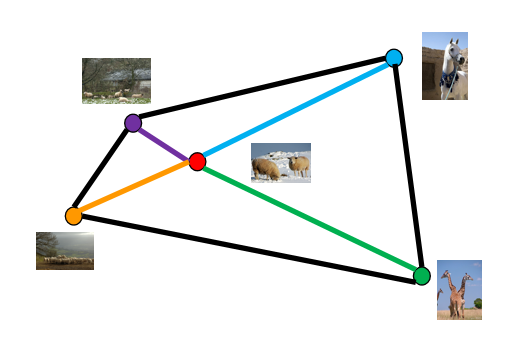} 
\begin{picture}(100,60)
\put(10,48){{\em\footnotesize [Up-down] Two sheep standing in the snow.}}
\put(10,24){{\em\footnotesize [GCN-LSTM] Two sheep standing and eating in the snow.}}
\put(10,36){{\em\footnotesize [AOA-net] Two sheep standing next to each other in the snow.}}
\put(10,12){{\em\footnotesize [M$^2$ Transformer] Two sheep stand in the snow and has nose to the ground.}}
\put(10,0){{\em\footnotesize [Dual-GCN+Trans+CL] Two sheep standing in the snow \textcolor{red}{with one looking for food.}}}
\end{picture}
\end{center}
\vspace{-0.2cm}
\caption{Visualization results compared with state-of-the-art methods. Note that we only plot top-4 images for simplicity.}
\label{fig:comp}
\vspace{-0.2cm}
\end{figure*}

\bfs{Visual Genome}~\cite{2017Visual} is a large dataset containing almost 108,000 images with densely annotated objects, attributes and relationships. We take almost 100,000 images for training, 5,000 images for validation and 5,000 images for testing as done in ~\cite{2018Bottom} and ~\cite{2018Exploring}. We observe that there are approximately 51,000 images in the visual genome are also found in the MS COCO dataset. To avoid contamination of our MS COCO validation and test sets, we make sure that no image will appear in both MS COCO datasets and the Visual genome dataset at the same time.
\bfsection{Evaluation Metrics} We adopt 4 widely used metrics to evaluate the performance of our proposed model on the image captioning task, including BLEU-n~\cite{2002A}, ROUGE~\cite{lin-2004-rouge}, METEOR~\cite{denkowski-lavie-2014-meteor} and CIDEr~\cite{2014CIDEr}. The BLEU-n metric evaluates the similarity between the generated caption and the reference caption. The choice of parameter $n$ refers to the reference caption in a different phrase. Therefore, a smaller number chosen as n means a better accuracy achieved by the generated caption; While a larger number of $n$ stands for a better semantic similarity between the generated caption and the reference. BLEU-1 and BLEU-4 are the most commonly used indicators in terms of accuracy and semantic similarity. Except for uni-gram precision, METEOR has adopted a weighted harmonic mean to evaluate the performance. ROUGE uses the longest captions between the generated caption and reference caption to measure their similarity at the caption level. CIDEr is a special one measuring human consensus for caption generation.



\vspace{-0.2cm}
\subsection{Implementation Detail}
Based on the Visual Genome dataset, we pretrain our object detector which is built upon the Faster-R-CNN~\cite{2017Faster} and enhanced by replacing its backbone with a resnet-101~\cite{2016Deep}.
For each input image, we firstly apply this detector to detect the object regions. Each region is represented by a 2048-dimensional vector $C=2048$. Then we select the top $O = 36$ regions with the highest confidences to represent the image. In the transformer decoder part, we use a 1000 dimensional word embedding to represent each word instead of using one-hot vectors. And we add the sinusoidal positional encodings into the word embedding. The number of transformer encoder layers and decoder layers are both set to 6. Similar to ~\cite{2017Attention}, the dimensionality of the layer and the number of the muti-head attention are set to 512 and 8, respectively. The number of similar images $K$ is set to 6. Toward the curriculum learning strategy, we calculate the average of BLEU-1 and BLEU-4 as the final score which turns out to be a more stable and realistic difficulty evaluation. We choose $M=8$ as the number of the datasets. 

 \begin{table*}[t!]
   \begin{center}
    \setlength{\tabcolsep}{3.5mm}
       \caption{Effectiveness comparison experiment on the MSCOCO dataset. Dual-GCN means the combination of GCN$_{obj}$ and GCN$_{img}$. $F_{obj}$ and $F_{img}$ indicate the input features fed into GCN$_{obj}$ and GCN$_{img}$, respectively. CL means Curriculum learning.}\label{Ablation}
       \vspace{-0.1in}
       \begin{tabular}{c|c|c|c|c|c|c|c}
       \hline
            & {\small BLEU}-1 &{\small BLEU}-2 & {\small BLEU}-3 & {\small BLEU}-4 & {\small METEOR} & {\small ROUGE} & {\small CIDEr}  \\
            \hline
        $F_{obj}$+Transformer   & 71.1     & 55.9  & 45.3   & 28.0  & 25.2 & 52.8 & 114.3 \\
        GCN$_{obj}$ + Transformer     & 80.5   & 64.7   & 49.2   & 37.6  & 27.9 & 57.4 & 125.8 \\
         $F_{obj}$+Transformer+CL    & 72.3     & 56.8  &  46.1  &  28.4 & 25.4 & 53.0 & 114.7 \\
         GCN$_{obj}$ + Transformer +CL  & 81.6     & 66.8   & 51.6   & 39.3  & 28.3   & 58.4  & {\bf 129.4} \\ 
         \hline
       $F_{img}$ + Transformer + CL &  80.2      &   64.1   &  49.0    &  37.4  &   27.4  &   57.0  & 124.0\\
       GCN$_{img}$ + Transformer +  CL    & 82.0   & 66.9   & 52.2   & {\bf 39.8}  & 29.6 & 59.4 & 129.0 \\
       \hline
      GCN$_{obj}$\&$F_{img}$ + Transformer + CL   &81.9      &67.3   &52.0    &39.5   &29.4  &59.6  & {\bf 129.4} \\
      GCN$_{img}$\&$F_{obj}$ + Transformer + CL   &82.0      & 67.3  & 52.3   &39.7   & {\bf 29.8} & 59.6 & 129.0  \\
      \hline
      Dual-GCN + Transformer   & 81.8     & 66.2  & 51.6   & 39.2  &28.7  & 58.9 &128.0  \\
      Dual-GCN + LSTM + CL   &  81.4    & 65.8  & 51.3   &  38.8 & 28.0 & 58.4  & 127.6 \\
      Dual-GCN + Transformer + CL & {\bf 82.2} & {\bf 67.6} &  {\bf 52.4} & 39.7 & 29.7 & {\bf 59.7} & 129.2
       \\
          \hline
       \end{tabular}
   \end{center}
        \vspace{-0.2in}  
\end{table*}

\subsection{Comparisons with State-of-the-Art}
\bfsection{Compared Methods} We compare our proposed DGCN with six state-of-the-art methods including (1) two attention based methods, {\em i.e.}, Up-down~\cite{2018Bottom} and  AOA Net~\cite{2019Attention}, (2) three graph based methods, {\em i.e.}, GCN-LSTM~\cite{2018Exploring}, GCN-LSTM+HIP~\cite{Yao2020Hierarchy} and SGAE~\cite{2018Auto}, and (3) one transformer based method, $M^2$ transformer~\cite{cornia2020m2}. 

\bfsection{Quantitative Comparison}
Table~\ref{tab:comp_with_state_of_the_art} shows the testing result of different models on the MS COCO dataset for image captioning task. Our proposed {\em Dual-GCN + Transformer + CL} model achieves the best results in all seven different evaluation metrics. Notably, it produces an 82.2 BLEU-1 score, which significantly outperforms other state-of-the-art models. This score is 2.0 higher than the traditional up-down method~\cite{2018Bottom}. Compared with the latest AOA method~\cite{2019Attention}, our model produces better BLEU-1 and CIDEr scores, which are 1.2 and 2.3 higher than AOA, respectively. We also observe that the self-attention mechanism in our transformer encoding and decoding process helps to improve the performance as compared to the traditional LSTM decoding used by GCN-LSTM~\cite{2018Exploring}. This is also justified by the fact that the score obtained by GCN-LSTM+HIP~\cite{Yao2020Hierarchy} is higher than that of the traditional LSTM based method \cite{2018Exploring}. However, the score of our model is higher than both GCN-LSTM~\cite{2018Exploring} and GCN-LSTM+HIP~\cite{Yao2020Hierarchy}. Finally, different from $M^2$ transformer~\cite{cornia2020m2} which uses memory blocks, our proposed {\em Dual-GCN + Transformer + CL}  model, which uses the Dual-GCN network, achieves the higher performance in terms of all the metrics. Obviously, these observations strongly demonstrate the efficacy of our proposed method. 


\bfsection{Qualitative Analysis} Figure~\ref{fig:comp} shows the qualitative results compared with four state-of-the-art methods. We can clearly see that our model produces a caption that correctly describes the content of the input image. Especially, the captions generated by our {\em Dual-GCN + Transformer + CL}  model are more detailed and authentic. 
\begin{figure*}[htbp]
\vspace{-0.3cm}
\begin{center}
\hspace{-5.5cm}\includegraphics[width=2.75cm, height=2.0cm]{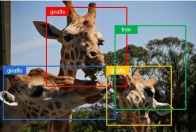} 
\includegraphics[width=2.75cm, height=2.0cm]{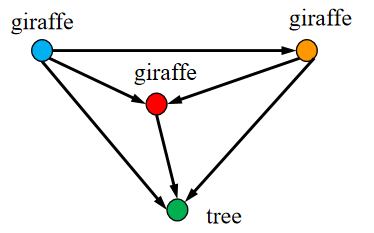} 
\hspace{0.25cm}\includegraphics[width=2.75cm, height=2.0cm]{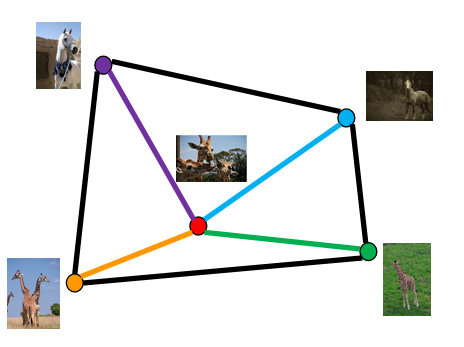} 
\begin{picture}(100,60)
\put(10,50){{\em\footnotesize [GT] Three giraffes are eating and grooming each oher in the zoo.}}
\put(10,40){{\em\footnotesize [GCN$_{img}$\&$F_{obj}$+Trans+CL] There are three giraffes are eating and standing.}}
\put(10,30){{\em\footnotesize [GCN$_{obj}$\&$F_{img}$+Trans+CL] Three giraffes are eating near a tree.}}
\put(10,20){{\em\footnotesize [Dual-GCN+Trans] Three giraffes standing and eating \textcolor{blue} {near a tree}.}}
\put(10,10){{\em\footnotesize [Dual-GCN+LSTM+CL] Three giraffes standing and eating next to each other.}}
\put(10,0){{\em\footnotesize [Dual-GCN+Trans+CL] Three giraffes standing and \textcolor{red}{eating near trees.}}}
\end{picture} \\
\vspace{0.05cm}
\hspace{-5.5cm}\includegraphics[width=2.75cm, height=2.0cm]{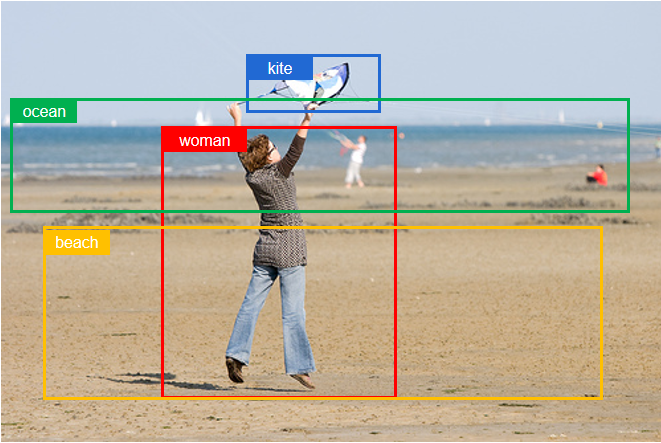} 
\includegraphics[width=2.75cm, height=2.0cm]{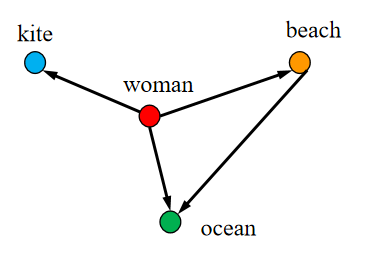} 
\hspace{0.25cm}\includegraphics[width=2.75cm, height=2.0cm]{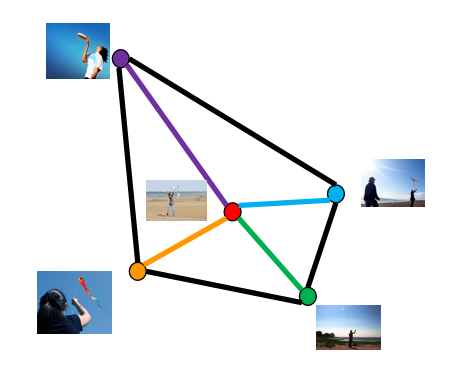} \begin{picture}(100,60)
\put(10,50){{\em\footnotesize [GT] A woman jumping up with a kite in her hands.}}
\put(10,40){{\em\footnotesize [GCN$_{img}$\&$F_{obj}$+Trans+CL] A person playing a kite on beach.}}
\put(10,30){{\em\footnotesize [GCN$_{obj}$\&$F_{img}$+Trans+CL] A woman holding a kite on beach.}}
\put(10,20){{\em\footnotesize [Dual-GCN+Trans] A woman jumping and playing a kite on beach.}}
\put(10,10){{\em\footnotesize [Dual-GCN+LSTM+CL] A person \textcolor{blue}{jumping and playing} a kite on beach.}}
\put(10,0){{\em\footnotesize [Dual-GCN+Trans+CL] A person \textcolor{red}{jumpling to hold} a kite on beach.}}
\end{picture} \\
\vspace{0.05cm}
\hspace{-5.5cm}\includegraphics[width=2.75cm, height=2.0cm]{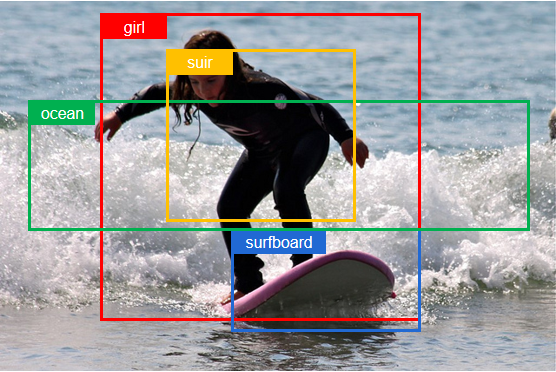} 
\includegraphics[width=2.75cm, height=2.0cm]{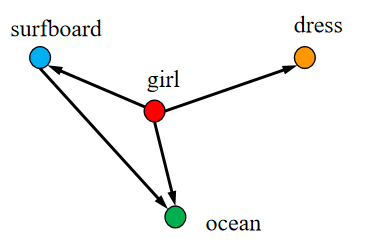} 
\hspace{0.25cm}\includegraphics[width=2.75cm, height=2.0cm]{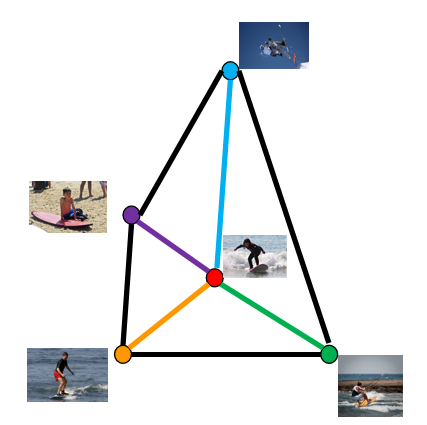} 
\begin{picture}(100,60)
\put(10,50){{\em\footnotesize [GT] A young girl tries to keep her balance on surfboard.}}
\put(10,40){{\em\footnotesize [GCN$_{img}$\&$F_{obj}$+Trans+CL] A girl playing on a surfboard on beach.}}
\put(10,30){{\em\footnotesize [GCN$_{obj}$\&$F_{img}$+Trans+CL] A girl \textcolor{blue}{playing on a surboard} in the ocean.}}
\put(10,20){{\em\footnotesize [Dual-GCN+Trans] A young girl standing on a surboard in the ocean.}}
\put(10,10){{\em\footnotesize [Dual-GCN+LSTM+CL] A girl riding a surfboard in the ocean.}}
\put(10,0){{\em\footnotesize [Dual-GCN+Trans+CL] A young girl \textcolor{red}{riding a surboard} in the ocean.}}
\end{picture} \\
\end{center}
\vspace{-0.2cm}
\caption{Visualization results of ablation experiments. Note that we only plot top-4 images for simplicity.}
\label{fig:ablation_study}
\vspace{-0.2in}
\end{figure*}
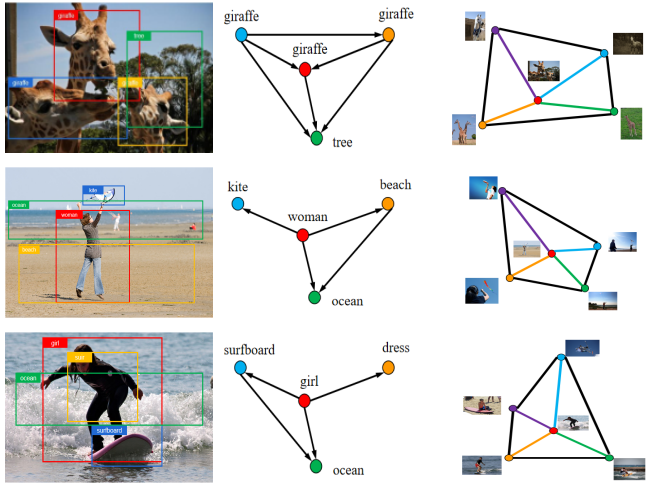
\vspace{-0.3cm}
\subsection{Ablation Study}
We conduct an ablation study to verify the effectiveness of different components in our method. To this end, we introduce several variants of our full model as shown in Table~\ref{Ablation}: (\romannumeral 1) $Transformer$ means only using the transformer and the proposed region embeddings to generate image captions. (\romannumeral 2) $F_{obj} $ means the object feature of the original image. (\romannumeral 3) $F_{img} $ means the mean pooling result of the $F_{obj} $. (\romannumeral 4) $GCN_{obj}$ means adopting the $GCN$ to process the region embeddings before feeding them into the transformer. (\romannumeral 5) $GCN_{img}$ means adopting the $GCN$ on $F_{img}$ before feeding them into the transformer. (\romannumeral 6) $CL$ means improving the previous model variant by using the Curriculum Learning strategy. (\romannumeral 7) $GCN_{obj}\&F_{img}$ and $GCN_{img}\&F_{obj}$ are two variants of our Dual-GCN in which both object-level and image-level visual features are used with only one GCN, either $GCN_{obj}$ or $GCN_{img}$. (\romannumeral 8) $LSTM$ means only using LSTM to replace the transformer to generate image captions. (\romannumeral 9) {\em Dual-GCN + Transformer + CL} is our full model and employs the proposed Dual-GCN module with transformer and curriculum learning to improve the quality of the final results further.
\bfsection{Effectiveness of Dual-GCN}
The proposed Dual-GCN network aims at enhancing the latent embedding. To verify the effectiveness of our Dual-GCN in improving the performance of image captioning. We have conducted a comparative experiment on whether the GCN network is used or not. The quantitative results are listed in Table \ref{Ablation}. When we use $GCN_{obj}\&F_{img}+Transformer+CL$, it shows that the BLEU-1 and BLEU-2 score reaches 81.9 and 67.3 respectively. Morever, when we use $GCN_{img}\&F_{obj}+Transformer+CL$, the score of BLUE-1 and BLEU-2 reaches 82.0 and 67.3. But until we added Dual-GCN in our model, the result turned out to be better: the BLEU-1 and BLEU-2 scores reach 82.2 and 67.6 respectively, which is higher than the model that only adopts one GCN network. This shows that our Dual-GCN does have a certain improvement effect on the model.

\bfsection{Effectiveness of Transformer}
The transformer has been used in image captioning only in recent years. To verify the difference between transformer and traditional LSTM, we added the contrast experiment between transformer and LSTM in our experiment. The quantitative results are listed in \ref{Ablation}. When using {\em Dual-GCN+LSTM+CL}, the BLEU-1 score reaches 81.4 and the CIDEr reaches 127.6. However, when we replace the original LSTM with the transformer, the result changes a lot: The BLEU-1 score reaches 82.2 and the CIDEr score reaches 129.2, which is 0.8 and 1.6 higher than {\em Dual-GCN+LSTM+CL} respectively. At the same time, the other of the evaluation scores also have been improved. Therefore, transformer does have a better effect than LSTM in our model.

\bfsection{Effectiveness of Curriculum Learning}
To verify the effectiveness of our proposed Curriculum Learning strategy, we have conducted a comparative experiment to train the transformer + GCN network either using curriculum learning or traditional training scheme. The experimental results are shown in Table \ref{Ablation}. We can find that Curriculum Learning  scheme introduces a specific improvement to our model. For example, without curriculum learning, the BLEU-1 score and BLEU-4 score only reached 81.4 and 38.8. When we adopt the curriculum learning to our
{\em Dual-GCN+transformer} model, it studies from easy to complex and the BLEU-1 and BLEU-4 reaches 82.2 and 39.7. This shows the effectiveness of our curriculum learning training method.

To make readers better understand our proposed method, we also provide visualization results for comparison in Figure~\ref{fig:ablation_study}. It shows the ground truth caption and different generated captions of our model in ablation experiments. For simplicity, here we only plot top-4 images.
\vspace{-0.3cm}
\subsection{Discussion}
\bfsection{Different Quantities of $K$ Similar Images}
In similarity evaluation part, we need to choose how many similar images should be used to generate the caption. To determine the value of parameter $K$, we compare the performance by using different values of parameter $K$. Table \ref{tab:K} shows the quantitative result. We can see that taking $K=6$ produces the best performance. This value is neither the smallest one nor the biggest one. This because smaller $K$ affects the accurate information of a single image, while more images cover more image features. When $K$ is set to 6, the score of BLEU-1 and BLEU-4 can reach the highest $82.2$ and $39.7$.
\begin{table}[ht!]
   \begin{center}
    \setlength{\tabcolsep}{1.5mm}
       \caption{The influence of different values of $K$.}    
       \vspace{-0.1in}    
       \begin{tabular}{c|c|c|c|c|c|c|c}
       \hline
            & B-1 & B-2 & B-3 & B-4 & METEOR & ROUGE & CIDEr  \\
            \hline
        K=3    & 80.2   & 64.8   & 49.4   & 37.2  & 27.4 & 57.0 & 124.7 \\
        K=4    & 80.7   & 65.0   & 50.2   & 37.4  & 27.9 & 57.4 & 124.9\\
        K=5    & 81.6   & 66.2   & 51.3   & 38.6  & 29.0 & 59.0 & 126.8 \\
        K=6    & {\bf 82.2 }  & {\bf 67.6}   & {\bf 52.4}   & {\bf 39.7}  & {\bf 29.7} & {\bf 59.7} & {\bf 129.2} \\
        K=7    & 82.0   & 67.1   & 52.2   & 39.4  & 29.4 & 59.5 & 128.9 \\
        K=8    & 81.8   & 66.7   & 51.7   & 38.9  & 29.2 & 59.2 & 127.6 \\
        K=9    & 82.0   & 67.0   & 51.9   & 39.1  & 29.4 & \bf{ 59.7} & 128.4 \\
            \hline
 \end{tabular}
 \vspace{-0.3in}    
 \label{tab:K}
   \end{center}
\end{table}

~\\
 
\bfsection{Different Quantities of $M$ Sub-datasets}
To make sure the specific influence of different parameter $M$ on the experimental results, we have tested our model using different integer values. 
Table \ref{tab:M} shows the quantitative result. The experimental results show the effect of increasing $M$ on the performance of our model. In a specific range, the scores obtained by our model increase with the increase of $M$. But the scores go down when the value of $M$ is bigger than a certain value. And our model obtains the best performance with $M=8$. 
We take $M=8$ in all other experiments.
\begin{table}[ht!]
 \vspace{0.1cm}
   \begin{center}
    \setlength{\tabcolsep}{1.4mm}
       \caption{The influence of different values of $M$.}    
       \vspace{-0.1in}     
       \begin{tabular}{c|c|c|c|c|c|c|c}
       \hline
            & B-1 & B-2 & B-3 & B-4 & METEOR & ROUGE & CIDEr  \\
            \hline
        M=5    & 81.7   & 66.5   & 51.4   & 38.7  & 28.9 & 59.0 & 128.2 \\
        M=6    & 82.1   & 67.0   & 52.3   & 39.4  & 29.4 & 59.6 & 128.9 \\
        M=7    & 81.8   & 66.6   & 51.6   & 39.0  & 29.0 & 59.1 & 123.2 \\
        M=8    & {\bf 82.2}   & {\bf 67.6}   & {\bf 52.4}   & {\bf 39.7}  & {\bf 29.7} & {\bf 59.7} & {\bf 129.2} \\
        M=9    & 81.6   & 66.4   & 51.2   & 38.4  & 28.7 & 58.7 & 127.4 \\
          \hline
       \end{tabular}
     \label{tab:M}   
   \end{center}
    \vspace{-0.2in}     
\end{table}

\begin{figure}
\captionsetup[subfigure]{justification=raggedright, singlelinecheck=false, labelformat=empty}
\begin{minipage}{0.23\linewidth}
\begin{subfigure}
    \centering
    \vspace{12pt}
    \includegraphics[width=\textwidth]{fig2.jpg}
    \makebox[1.0\textwidth]{\footnotesize Input Image}
\end{subfigure} 
\end{minipage}
\begin{minipage}{0.70\linewidth}
    \includegraphics[height=1cm,width=0.243\textwidth]{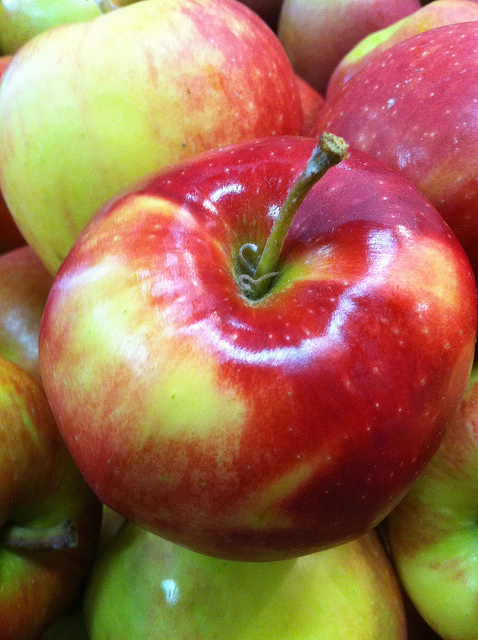}
    \includegraphics[height=1cm,width=0.243\textwidth]{fig3.jpg}
    \includegraphics[height=1cm,width=0.243\textwidth]{fig4.jpg}
    \includegraphics[height=1cm,width=0.243\textwidth]{fig5.jpg}\\
    \includegraphics[height=1cm,width=0.243\textwidth]{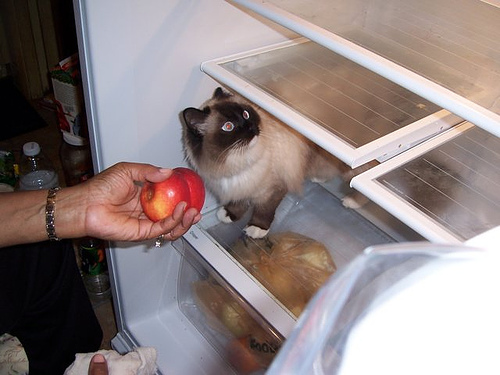}
    \includegraphics[height=1cm,width=0.243\textwidth]{fig7.jpg}
    \includegraphics[height=1cm,width=0.243\textwidth]{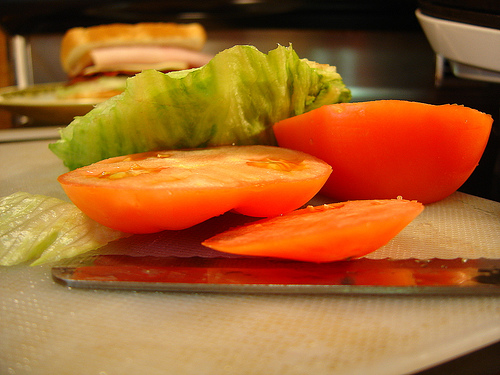}
    \includegraphics[height=1cm,width=0.243\textwidth]{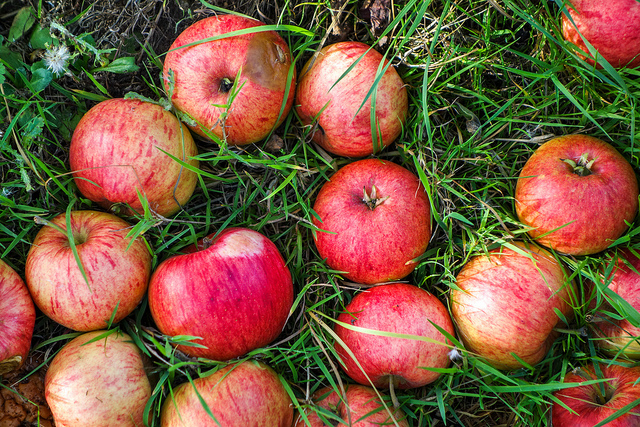}
\end{minipage}%
\vspace{-0.42cm}
\begin{picture}(100,20)
\put(-60,10){[GT] {\em A bunch of red apples and yellow oranges in a closeup.}}
\put(-60,0){[Output] {\em A red apple \textcolor{blue}{sitting on top of two apples}.}}
\end{picture}
\caption{ The result of similar image selection failure. When guiding the generation of apple, the model chooses the man in red clothes as the guide image with high similarity.}\label{limitation}
\vspace{-0.2in}
\end{figure}

\bfsection{Limitation and Failure Cases}
To clarify, the measurement of similarity between different images largely depends on the $\mathcal{L}_2$ distance of the image feature vector. Therefore, when our algorithm wrongly selects images to guide the image captioning of the original image, it will have some negative effects. Figure~\ref{limitation} shows the visualization of negative examples. 

\section{Conclusion}
We propose a novel image captioning model in a combination of Dual-GCN and Transformer with Curriculum Learning as training strategy. The visual feature encoded by an object-level GCN and an image-level GCN is designed to incorporate both local and global visual encoding. The transformer decoder is able to understand the extracted visual features to generate reasonable caption results. With a cross-review mechanism to determine the difficulty of the dataset, we take curriculum learning as the training strategy to ensure that our proposed model is trained in a manner from easy to hard. Experiments conducted on the MS COCO dataset have strongly confirmed the efficacy of our proposed method.

\section*{Acknowledgement}
This work was partly supported by the Key Technological Innovation Projects of Hubei Province (2018AAA062), NSFC (NO. 61972298), and CAAI-Huawei MindSpore
Open Fund.

\bibliographystyle{ACM-Reference-Format}
\bibliography{DGCN_MM}

\appendix

\end{document}